\documentclass[10pt,twocolumn,letterpaper]{article}

\usepackage{cvpr}
\usepackage{times}
\usepackage{epsfig}
\usepackage{graphicx}
\usepackage{amsmath}
\usepackage{amssymb}

\usepackage{comment}

\usepackage{xfrac}
\usepackage{subcaption}
\usepackage[export]{adjustbox}
\usepackage[table]{xcolor}
\usepackage{color}
\usepackage{rotating}
\usepackage{array}
\usepackage{siunitx}
\usepackage{units}

\definecolor{myblue}{rgb}{.1,0.3,0.9}
\definecolor{rowblue}{RGB}{220,230,240}
\usepackage{booktabs}
\usepackage{tablefootnote}
\usepackage{colortbl}
\usepackage{tabularx}
\usepackage{multirow}

\usepackage[pagebackref=true,breaklinks=true,letterpaper=true,colorlinks,bookmarks=false]{hyperref}

\cvprfinalcopy 

\def\cvprPaperID{5436} 

\begin{document}

\title{\vspace{-30pt}ZeroScatter: Domain Transfer for Long Distance Imaging and Vision\\through Scattering Media\vspace{-10pt}}
\author{Zheng Shi$^1$\thanks{indicates equal contribution.} \qquad Ethan Tseng$^1$\footnotemark[1] \qquad Mario Bijelic$^2$\footnotemark[1] \qquad Werner Ritter$^2$ \qquad Felix Heide$^{1,3}$\vspace{5pt}\\
$^1$Princeton University \qquad $^2$Mercedes-Benz AG \qquad $^3$Algolux\\
}

\maketitle


\definecolor{Gray}{rgb}{0.5,0.5,0.5}
\definecolor{darkblue}{rgb}{0,0,0.7}
\definecolor{orange}{rgb}{1,.5,0} 
\definecolor{red}{rgb}{1,0,0} 

\newcommand{\heading}[1]{\noindent\textbf{#1}}
\newcommand{\note}[1]{{\em{\textcolor{orange}{#1}}}}
\newcommand{\todo}[1]{{\textcolor{red}{\bf{TODO: #1}}}}
\newcommand{\comments}[1]{{\em{\textcolor{orange}{#1}}}}
\newcommand{\changed}[1]{#1}
\newcommand{\place}[1]{ \begin{itemize}\item\textcolor{darkblue}{#1}\end{itemize}}
\newcommand{\de}{\mathrm{d}}

\newcommand{\normlzd}[1]{{#1}^{\textrm{aligned}}}

\newcommand{\ttime}{\tau}               
\newcommand{\x}{\Vect{x}}               
\newcommand{\z}{z}               

\newcommand{\npixels}{n}               
\newcommand{\ntime}{t}               

\newcommand{\illfunc}     {g}
\newcommand{\pathfunc}     {s}
\newcommand{\camfunc}     {f}

\newcommand{\irradiance}{E}
\newcommand{\exposure}{b}
\newcommand{\pmdfunc}{f}                
\newcommand{\lightfunc}{g}              
\newcommand{\period}{T}                 
\newcommand{\freqm}{\omega}                
\newcommand{\illphase}{\rho}             
\newcommand{\sensphase}{\psi}             
\newcommand{\pmdphase}{\phi}            
\newcommand{\omphi}{{\omega,\phi}}      
\newcommand{\numperiod}{N}              
\newcommand{\att}{\alpha}               
\newcommand{\pathspace}{{\mathcal{P}}}  

\newcommand{\atan}{\operatorname{atan}}

\newcommand{\Fourier}{\mathfrak{{F}}}         
\newcommand{\conv}     {\otimes}
\newcommand{\corr}     {\star}
\newcommand{\Mat}[1]    {{\ensuremath{\mathbf{\uppercase{#1}}}}} 
\newcommand{\Vect}[1]   {{\ensuremath{\mathbf{\lowercase{#1}}}}} 
\newcommand{\Id}				{\mathbb{I}} 
\newcommand{\Diag}[1] 	{\operatorname{diag}\left({ #1 }\right)} 
\newcommand{\Opt}[1] 	  {{#1}_{\text{opt}}} 
\newcommand{\CC}[1]			{{#1}^{*}} 
\newcommand{\Op}[1]     {\Mat{#1}} 
\newcommand{\minimize}[1] {\underset{{#1}}{\operatorname{argmin}} \: \: } 
\newcommand{\maximize}[1] {\underset{{#1}}{\operatorname{argmax}} \: \: } 
\newcommand{\grad}      {\nabla}

\newcommand{\Basis}{\Mat{H}}         		
\newcommand{\Corr}{\Mat{C}}             
\newcommand{\correlem}{\bold{c}}             
\newcommand{\meas}{\Vect{b}}            
\newcommand{\Meas}{\Mat{B}}            
\newcommand{\MeasNormalized}{\Mat{B}^{\textrm{new}}}            
\newcommand{\Img}{H}                    
\newcommand{\img}{\Vect{h}}             
\newcommand{\latentresponse}{\alpha}

\begin{abstract}
Adverse weather conditions, including snow, rain, and fog, pose a major challenge for both human and computer vision. Handling these environmental conditions is essential for safe decision making, especially in autonomous vehicles, robotics, and drones. Most of today's supervised imaging and vision approaches, however, rely on training data collected in the real world that is biased towards good weather conditions, with dense fog, snow, and heavy rain as outliers in these datasets. Without training data, let alone paired data, existing autonomous vehicles often limit themselves to good conditions and stop when dense fog or snow is detected. In this work, we tackle the lack of supervised training data by combining synthetic and indirect supervision. We present ZeroScatter, a domain transfer method for converting RGB-only captures taken in adverse weather into clear daytime scenes. ZeroScatter exploits model-based, temporal, multi-view, multi-modal, and adversarial cues in a joint fashion, allowing us to train on unpaired, biased data. We assess the proposed method on in-the-wild captures, and the proposed method outperforms existing monocular descattering approaches by 2.8 dB PSNR on controlled fog chamber measurements.
\end{abstract}
	
\vspace{-8pt}
\section{Introduction}
\begin{figure*}[t!]
	\centering
	\includegraphics[width=\textwidth]{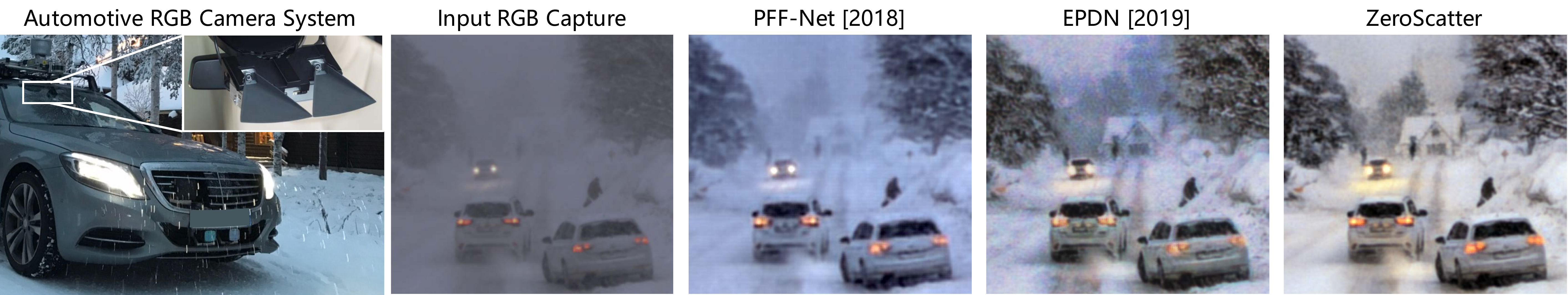}
	\vspace*{-14pt}
	\caption{Scattering stemming from snow, rain, or fog significantly reduces the perceptible quality of RGB captures and impact downstream computer vision tasks such as object detection. The proposed method, which we dub ``ZeroScatter'', reliably removes these scattering effects for unseen automotive scenes.}
	\label{fig:teaser}
    \vspace*{-12pt}
\end{figure*}

In the presence of a scattering medium, such as fog or snow, photons no longer propagate along a straight path but instead are redirected by particles, potentially many times, until arriving at the camera. This includes forward scattered light emitted from sources in the scene, e.g., an oncoming vehicle headlight, captured as a passive component by an RGB camera or human eye, and backward scattering observed when actively illuminating the scene, e.g., in automotive lidar or with the ego-vehicle headlights. While adverse weather conditions that include severe scattering are heavily underrepresented in existing training and evaluation datasets~\cite{sun2020scalability,geiger2012we,Cordts2016}, these rare scenarios are a significant contributing factor for fatal automotive accidents~\cite{DrivingBlind2015}, as a direct result of vision impairment for human drivers.

Supervised imaging and vision approaches are also fundamentally limited in adverse weather conditions. Adverse weather conditions follow a long-tail distribution where such environments are rarely encountered during day-to-day driving, making data collection, training, and evaluation challenging~\cite{Oldenborgh2009OnTR}. As a result, critical computer vision tasks such as object detection and tracking are often trained on clear day inputs and fail to generalize when the input scene is perturbed by adverse effects from scattering media. Even if adverse weather data is available, the scattering media would still affect the quality of human annotations used for supervision. Furthermore, supervised dehazing and defogging methods are restricted by the difficulty of acquiring paired perturbed and clear data, which is infeasible due to the dynamic nature of real-world automotive scenes. As such, supervised training on real-world data has been a fundamental challenge for imaging and vision in harsh weather conditions. To tackle this problem, existing approaches attempt to solve a domain transfer problem using simulated scattering media~\cite{mei2018pffn,mondal2018imagedb,sakaridis2018semantic,halder2019physics}. However, these simulation models do not adequately simulate the effects that are observed in the wild. Unsupervised learning approaches have demonstrated impressive ability for image domain transfer but remain restricted to a single domain, e.g. faces, and small image resolutions~\cite{zhu2017unpaired,Karras2019stylegan2}.

Researchers have also adopted alternative sensing modalities beyond conventional intensity imaging, e.g. lidar and radar, in robotic and automotive applications. However, they do not offer a solution in backscatter-limited weather scenarios. Specifically, pulsed lidar sensors that record the round-trip time of the first response fail to extract meaningful scene surfaces in severe snow and fog, fundamentally limited by backscatter~\cite{Bijelic_2020_STF}, and indeed trail the performance of RGB stereo depth methods~\cite{gruber2019pixel} in dense fog. While the mm-wavelengths of radar systems penetrate dense fog, existing radar systems are limited to low angular resolution, and hence do not allow for scene understanding tasks beyond the detection and tracking of objects with a large radar cross-section~\cite{RadarAngularResolution}. At the same time, RGB intensity cameras have become a ubiquitous sensor technology because of their low-cost and high spatial resolutions up to 250~MPix in modern commodity sensors~\cite{Samsung250MPix}, deployed across application domains from miniature smartphone cameras to automotive imaging systems. As such, in this work, we address the task of imaging through scattering media using conventional RGB cameras.

We tackle this challenge by proposing ZeroScatter, a novel domain transfer method that converts RGB images corrupted by adverse weather effects into clear day scenes. To do this, we exploit a variety of training signals in order to achieve robust descattering performance on real-world examples. First, we employ a synthetic weather model using cycle consistency training. Second, we employ temporal and multi-view consistency to ensure stable model performance and to eliminate spurious adverse weather effects such as snowflakes, leveraging an adverse weather dataset~\cite{Bijelic_2020_STF}. Third, we employ multi-modal supervision using auxiliary data acquired by gated imagers~\cite{gruber2019gated2depth}. Gated imaging is an emerging time-of-flight imaging technology that records photons with specific return times which allows it to image objects at select distances. This imaging modality is less susceptible to path lengths and provides higher contrast training signal for ZeroScatter. All of these training cues enable ZeroScatter to reliably reconstruct RGB captures that have been corrupted by adverse weather. For quantitative evaluation, we evaluated ZeroScatter on scenes with synthetically generated and laboratory generated adverse weather where we demonstrate $\SI{2.8}{dB}$ PSNR improvement over state-of-the-art methods.

Specifically, we make the following contributions:
\vspace{-5pt}
\begin{itemize}	
	\itemsep-0.3em
	\item We propose a novel domain adaptation method which we call ZeroScatter for eliminating scattering media from conventional RGB captures,  operating at real-time frame rates of 20 FPS.
	\item We employ a novel combination of synthetic and real-world data to train ZeroScatter with unpaired, biased datasets. To this end, we incorporate model-based cues jointly with multi-modal, multi-view, temporal and adversarial cues.
	\item In addition to qualitative improvements on real-world captures, we outperform state-of-the-art methods in controlled fog-chamber evaluation. Our method also outperforms state-of-the-art object detection in harsh weather at long distances.
\end{itemize}

\section{Related Work}
\vspace{-2pt}
\paragraph{Descattering} A variety of image descattering techniques have been proposed in recent years. Several works have been proposed for single dedicated tasks such as dehazing~\cite{cai2016dehazenet, Li2017,mei2018pffn,mondal2018imagedb,EnhancedPix2Pix}, removing rain~\cite{Hu_2019_CVPR,Clearing_the_skies, GANImageDeraining2017Zhang,Pu2018,Zhang2018b,zheng_2020_ECCV,Wang_2020_CVPRModelRain}, removing snow~\cite{liu2018desnownet}, and translating night to day~\cite{zheng_2020_ECCV}.

Earlier descattering approaches that employed convolutional neural networks (CNNs)~\cite{cai2016dehazenet,mondal2018imagedb} learned the scattering effects as a residual image by separately estimating the airlight and the transmission. However, this disjoint learning approach can amplify prediction errors. Li et al.~\cite{Li2017} proposes to learn both parameters in an end-to-end fashion by inverting the image formation model. Similar approaches have been proposed for removing rain~\cite{Clearing_the_skies,Wang_2020_CVPRModelRain}. These methods demonstrate strong performance through their explicit image formation models., but are difficult to apply to other adverse weather types. Recent methods~\cite{mei2018pffn, Hu_2019_CVPR} directly learn the desired descattering without a prescribed image formation model. These methods are trained entirely using synthetically simulated weather conditions, and therefore struggle with real-world scenes.

\vspace{-5pt}
\paragraph{Domain Adaptation} The recent development of GAN architectures~\cite{Zhang2018cDCPDN, EnhancedPix2Pix} has demonstrated impressive results for image translation. However, most of these methods require paired simulated data consisting of full pixel-wise ground truth images for supervised training.

Methods that do not require paired ground truth~\cite{Engin2018,Zhao2019,Yang_2020_CVPRCylceRain} are based on CycleGAN~\cite{zhu2017unpaired}. While this allows for better training stability, it is difficult to learn both directions of the cycle, specifically the descattering and re-scattering processes. We alleviate these limitations for ZeroScatter by employing a novel cycle training approach where we train the descatterer but utilize a fixed adverse weather simulator for the reverse direction of the cycle. Furthermore, our use of temporal, multi-view, and multi-modal supervision improves ZeroScatter's generalization to real-world inputs over methods that do not utilize additional cues.

\vspace{-5pt}
\paragraph{Weather Simulation and Datasets} Adverse weather simulation techniques have been developed for snowfall~\cite{liu2018desnownet}, rainfall~\cite{hasirlioglu2019rainsimulation,halder2019physics}, blur \cite{kupyn2018deblurgan}, fog~\cite{Li2017a,Galdran2018,Sakaridis2018}, night driving \cite{sakaridis2019guided,Liu2017}, and raindrops on the windshield \cite{Bernuth2018}. Most datasets~\cite{Sakaridis2018,tarel2010improved,Ancuti2016,Zhang2017cHazeRD,Li2017a,Li2019Reside} are based on Koschmieder's physical model \cite{Koschmieder1924}. These techniques overlay clear weather images with one type of adverse weather perturbation to create paired examples for supervised training. Very few datasets contain real-world adverse weather scenes~\cite{Ancuti2018,ancuti2018haze,Li2019Reside,gruber2019pixel}. RESIDE~\cite{Li2019Reside} contains 4322 real foggy scenes obtained from the internet, along with their annotated object detection labels to enable task-driven dehazing. The O-HAZE~\cite{ancuti2018haze} and I-HAZE~\cite{Ancuti2018} datasets contain real outdoor and indoor hazy scenes respectively which were generated with professional haze machines. However, the datasets are very small with only 45 outdoor and 35 indoor image pairs. Gruber et al.~\cite{gruber2019pixel} provides a recent depth benchmark dataset with four scenes under different conditions that, as such, is too small for training purposes. In order to provide a variety of training cues for ZeroScatter, we utilize an adverse weather dataset containing real-world automotive captures from northern Europe~\cite{Bijelic_2020_STF}. In addition to RGB captures, the dataset consists of multi-modal data in the form of gated images~\cite{grauer2014active}, multi-view stereo data, and temporal sequences.

\section{Domain Transfer with ZeroScatter}
\begin{figure*}[t!]
	\centering
	\includegraphics[width=\textwidth]{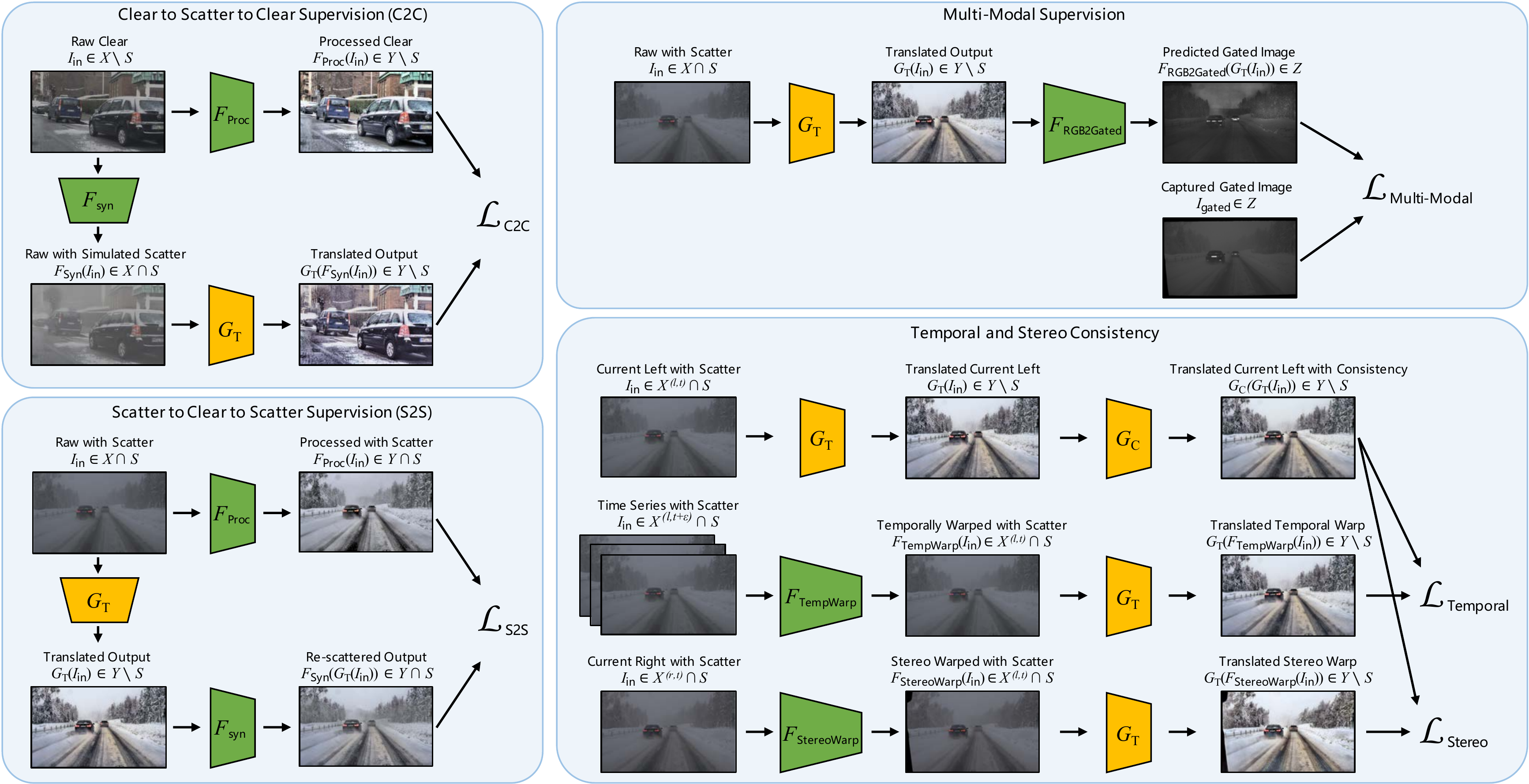}
	\vspace*{-16pt}
	\caption{Overview of the proposed method. We train our generator using a novel combination of training cues that promote high-contrast, scatter-free, jitter-free results on unseen real-world scenes. We employ model-based supervision using cycle training which is facilitated by a robust adverse weather model, multi-modal supervision in the form of gated images for training on real heavy weather scenes, and consistency supervision in the form of temporal and stereo losses.}
	\label{fig:pipeline}
	\vspace*{-10pt}
\end{figure*}

\subsection{Formulation}

\setlength{\belowdisplayskip}{5pt} \setlength{\belowdisplayshortskip}{5pt}
\setlength{\abovedisplayskip}{5pt} \setlength{\abovedisplayshortskip}{5pt}

To train a reconstruction network $G$ without supervised training data available, we employ cues from adverse weather simulation,  multi-modal cues that other sensors can provide, multi-view cues, and temporal consistency cues. Specifically, let $X$ be the domain of raw RGB images, $Y$ be the (unpaired) domain of processed daytime RGB images, and $S$ be the (unpaired) domain of RGB images with scattering present. We train the mapping $G : X \cap S \rightarrow Y \setminus S$, which itself is composed of a translation block $G_\text{T} : X \cap S \rightarrow Y \setminus S$ for image domain transfer and a consistency block $G_\text{C} : Y \setminus S \rightarrow Y \setminus S$ for minimizing temporal and spatial jitter. As illustrated in Figure~\ref{fig:pipeline}, we employ several auxiliary mapping functions to facilitate our learning scheme.

The model-based learning cycles utilize a user-defined ISP processing function $F_\text{Proc} : X \rightarrow Y$ and an adverse weather simulator $F_\text{Syn} : S^c \rightarrow S$. These mappings enable two training cycles, one involving clear daytime images, which we call ``Clear to Scatter to Clear'':
\begin{equation}
I_\text{in} \rightarrow F_\text{Syn}(I_\text{in}) \rightarrow G_\text{T}(F_\text{Syn}(I_\text{in})) \approx F_\text{Proc}(I_\text{in}),
\end{equation}
where $I_\text{in} \in X \setminus S$ is clear daytime images; and another involving scatter corrupted daytime images which we call ``Scatter to Clear to Scatter'':
\begin{equation}
I_\text{in} \rightarrow G_\text{T}(I_\text{in}) \rightarrow F_\text{Syn}(G_\text{T}(I_\text{in})) \approx F_\text{Proc}(I_\text{in}),
\end{equation}
where $I_\text{in} \in X \cap S$ is scatter corrupted daytime images. 

Indirect supervision with multi-modal data is performed using gated images, as it is less affected by scatters. We pre-train a neural network $F_\text{RGB2Gated}: Y \setminus S \rightarrow Z$ for inferring gated images $Z$ from processed clear daytime scenes. We then use it with the real captured gated images $I_\text{gated}$:
\begin{equation}
I_\text{in} \rightarrow F_\text{RGB2Gated}(G_\text{T}(I_\text{in})) \approx I_\text{gated}
\end{equation}
where $I_\text{in} \in X \cap S$ is scatter corrupted daytime images. 
Lastly, we utilize temporal and multi-view data as learning cues. This indirect supervision is facilitated by a temporal warper $F_\text{TempWarp} : X^{(t+\epsilon)} \rightarrow X^{(t)}$ which warps temporally adjacent frames to the current frame and a stereo warper $F_\text{StereoWarp} : X^{(r)} \rightarrow X^{(l)}$ which warps the right stereo image $X^{(r)} = X \cap R$ onto the left viewpoint $X^{(l)} = X \cap L$. We feed the warped images in addition to the current left capture through $G_\text{T}$ and then we train $G_\text{C}$ to complete the following training paths:
\begin{equation}
I^{(l,t)}_\text{in} \rightarrow G_\text{C}(G_\text{T}(I^{(l,t)}_\text{in})) \approx G_\text{T}(F_\text{TempWarp}(I^{(l,t+\epsilon)}_\text{in})),
\end{equation}
and
\begin{equation}
I^{(l,t)}_\text{in} \rightarrow G_\text{C}(G_\text{T}(I^{(l,t)}_\text{in})) \approx G_\text{T}(F_\text{StereoWarp}(I^{(r,t)}_\text{in})).
\end{equation}
In the following, we first describe each of these training components in more detail before discussing the generator architecture we employ for $G_\text{C}(G_\text{T}(\cdot))$.

\subsection{Model-Based Synthetic Supervision}
Our model-based training scheme has two training cycles called ``Clear to Scatter to Clear'' (C2C) and ``Scatter to Clear to Scatter'' (S2S). We employ a fixed adverse weather simulator $F_\text{Syn} : S^c \rightarrow S$ that applies simulated adverse weather to RGB images, based on haze estimation following Koschmieder's model~\cite{Koschmieder1924} with several modifications that promote generalization to real-world scenes. This ensures ZeroScatter is able to handle various-intensity and depth-dependent scatter effects. As many computer vision applications consume ISP processed images instead of raw camera captures, we also employ a post-processing function $F_\text{Proc} : X \rightarrow Y$. This function can be arbitrarily defined by the user, for this work we define $F_\text{Proc}$ to be a raw capture to daytime RGB mapping. These two functions are applied in a cyclic manner to the output of the generator translation block $G_\text{T} : X \cap S \rightarrow Y \setminus S$ as shown in Figure~\ref{fig:pipeline}. For more detail on $F_\text{Syn}$ and $F_\text{Proc}$ please refer to the Supplemental Document.

Our model-based supervision aims to minimize 
\begin{equation}
\mathcal{L}_\text{Model} = \mathcal{L}_\text{C2C} + \mathcal{L}_\text{S2S}.
\end{equation}
For the C2C cycle we compute the loss using the input clear weather image $I_\text{in}\in X \setminus S$ : 
\begin{equation}
\mathcal{L}_\text{C2C} = (\mathcal{L}_1 + \mathcal{L}_\text{perc} + \mathcal{L}_\text{grad} + \mathcal{L}_\text{adv})( I_\text{T}, I_\text{target}),
\end{equation}
where $I_\text{T} = G_\text{T}(F_\text{Syn}(I_\text{in}))$ and  $I_\text{target} = F_\text{Proc}(I_\text{in})$ is the processed target image, $\mathcal{L}_1$ is the Mean Absolute Error loss, $\mathcal{L}_\text{perc}$ is a VGG-19 based perceptual loss~\cite{johnson2016perceptual}, $\mathcal{L}_\text{grad}$ is an image gradient loss, and $\mathcal{L}_\text{adv}$ is a GAN based adversarial loss~\cite{goodfellow2014generative}. 

For the S2S cycle we compute the loss using the input adverse weather image $I_\text{in} \in X \cap S$ as
\begin{equation}
\mathcal{L}_\text{S2S} = \mathcal{L}_\text{adv}(F_\text{Proc}(I_\text{in}), F_\text{Syn}(G_\text{T}(I_\text{in}))).
\end{equation}
Since there are a wide variety of plausible adverse scatter effects, we avoid using $\mathcal{L}_1$, $\mathcal{L}_\text{grad}$ and $\mathcal{L}_\text{perc}$ in the S2S cycle and instead use only an adversarial loss.

\subsection{Multi-Modal Indirect Supervision}

We employ a multi-modal indirect supervision approach to facilitate training on data captured in-the-wild, which makes use of emerging gated imagers \cite{grauer2014active,gruber2019gated2depth,Bijelic_2020_STF} that uses active flash illumination to acquire high contrast images by temporally gating out scattering components.

As such, gated images are less affected by adverse weather than RGB cameras \cite{Bijelic2018}. However, they cannot be directly used for training supervision due to the domain shift between gated images and RGB images, e.g. gated images lack color information, see Fig.~\ref{fig:pipeline}. To overcome this domain shift, we train an RGB2Gated network $F_\text{RGB2Gated} : Y \setminus S \rightarrow Z$, where $Z$ is the domain of gated images. This network predicts the gated image corresponding to a processed clear day RGB capture. By training our RGB2Gated network only on clear day images, we teach the network to predict the gated image in the absence of scattering media. We apply $F_\text{RGB2Gated}$ to the RGB output of $G_\text{T}$, which then allows us to compute a loss with respect to the actual gated image. As a result, our gated supervision loss encourages our generator to remove adverse weather effects to match the underlying image with scattering removed. For details on the RGB2Gated network architecture and training procedure please see the Supplemental Document.

During training we apply $F_\text{RGB2Gated}$ to $I_\text{T} = G_\text{T}(I_\text{in})$, $I_\text{in} \in X\cap S$, and compare the resulting image $I'_\text{gated}$ to the corresponding real gated image $I_\text{gated}$. To filter out areas that contain insufficient information due to extreme long distance and overly strong reflections from retroreflectors, we apply a mask $M_\text{ent}$ based on the local entropy of the real gated capture. The multi-modal supervision loss is expressed as
\begin{equation}
\mathcal{L}_\text{Multi-Modal} = \mathcal{L}_\text{perc}(M_\text{ent}\odot I_\text{gated}, M_\text{ent}\odot I'_\text{gated}),
\end{equation}
where $\odot$ is point-wise multiplication.

We emphasize that we only use gated images for training supervision and that the generator only requires RGB inputs at test time. Our multi-modal loss provides a better training signal for the proposed ZeroScatter method but does not require the specialized gated imaging system at test time.

\subsection{Temporal and Stereo Consistency}
We employ an indirect consistency supervision to ensure a temporally and stereo consistent output. To do this, we align the multi-view and temporal outputs of our network with respect to the current left viewpoint. For stereo rectification, this is done by employing a depth-based warp $F_\text{StereoWarp} : X^{(r)} \rightarrow X^{(l)}$ which maps the right viewpoint images onto the left viewpoint images. For temporal alignment we apply an optical flow warp $F_\text{TempWarp} : X^{(t+\epsilon)} \rightarrow X^{(t)}$ to determine a warped current image from a temporally adjacent frame. For details on the warping procedures please refer to the Supplemental Document.

In addition to temporal and stereo consistency losses during training, we employ a consistency block $G_\text{C} : Y \setminus S \rightarrow Y \setminus S$ as a downstream network after the translation block to achieve high quality consistent outputs. Directly applying the consistency losses to a single-stage network produces inferior results as the single-stage network struggles to remove both fine scattering effects, such as haze and coarse scattering effects such as snowflakes, in addition to other jitters such as sensor noise. We train $G_\text{C}$ using the consistency losses while $G_\text{T}$ focuses on the other losses previously described. See our ablation comparison in Section~\ref{sec:ablation} for the benefits of our two-stage sequential network.

Putting everything together, we train the consistency block $G_\text{C}$ to minimize the following consistency loss:
\begin{equation}
\mathcal{L}_\text{Consistency} = \mathcal{L}_\text{Temp} + \mathcal{L}_\text{Stereo}.
\end{equation}
The temporal loss component is computed as 
\begin{equation}
\begin{split}
\mathcal{L}_\text{Temp} &= (\mathcal{L}_\text{1} + \mathcal{L}_\text{perc})(G_\text{C}(G_\text{T}(I_\text{in})), G_\text{T}(I'_\text{in})),
\end{split}
\end{equation}
where
\begin{equation}
I'_\text{in} = M_\text{Temp}(F_\text{TempWarp}(I^{(t-1)}_\text{in}),F_\text{TempWarp}(I^{(t+1)}_\text{in}))
\end{equation}
is the warped current input computed from temporally adjacent frames $I^{(t-1)}_\text{in} \in X^{(t-1)}$ and $I^{(t+1)}_\text{in} \in X^{(t+1)}$, and $M_\text{Temp}$ is a visibility mask that merges the two warped temporally adjacent frames by recovering out-of-view pixels and occlusions, see Supplemental Document for details.

The stereo loss component is computed as 
\begin{equation}
\begin{split}
\mathcal{L}_\text{Stereo} &= M_\text{Stereo} \odot \mathcal{L}_\text{1}(G_\text{C}(G_\text{T}(I_\text{in})), G_\text{T}(I'_\text{in})),
\end{split}
\end{equation}
where $I'_\text{in} = F_\text{StereoWarp}(I^{(r)}_\text{in})$ is the warped right stereo image, and $M_\text{Stereo} = \exp(-\alpha \mathcal{L}_\text{1} (I_\text{in}, I'_\text{in}))$ is a visibility mask calculated from the warping error between the left input and the warped right input, and we empirically set $\alpha = 10$.

\begin{figure}[t!]
	\centering
	\includegraphics[width=0.46\textwidth]{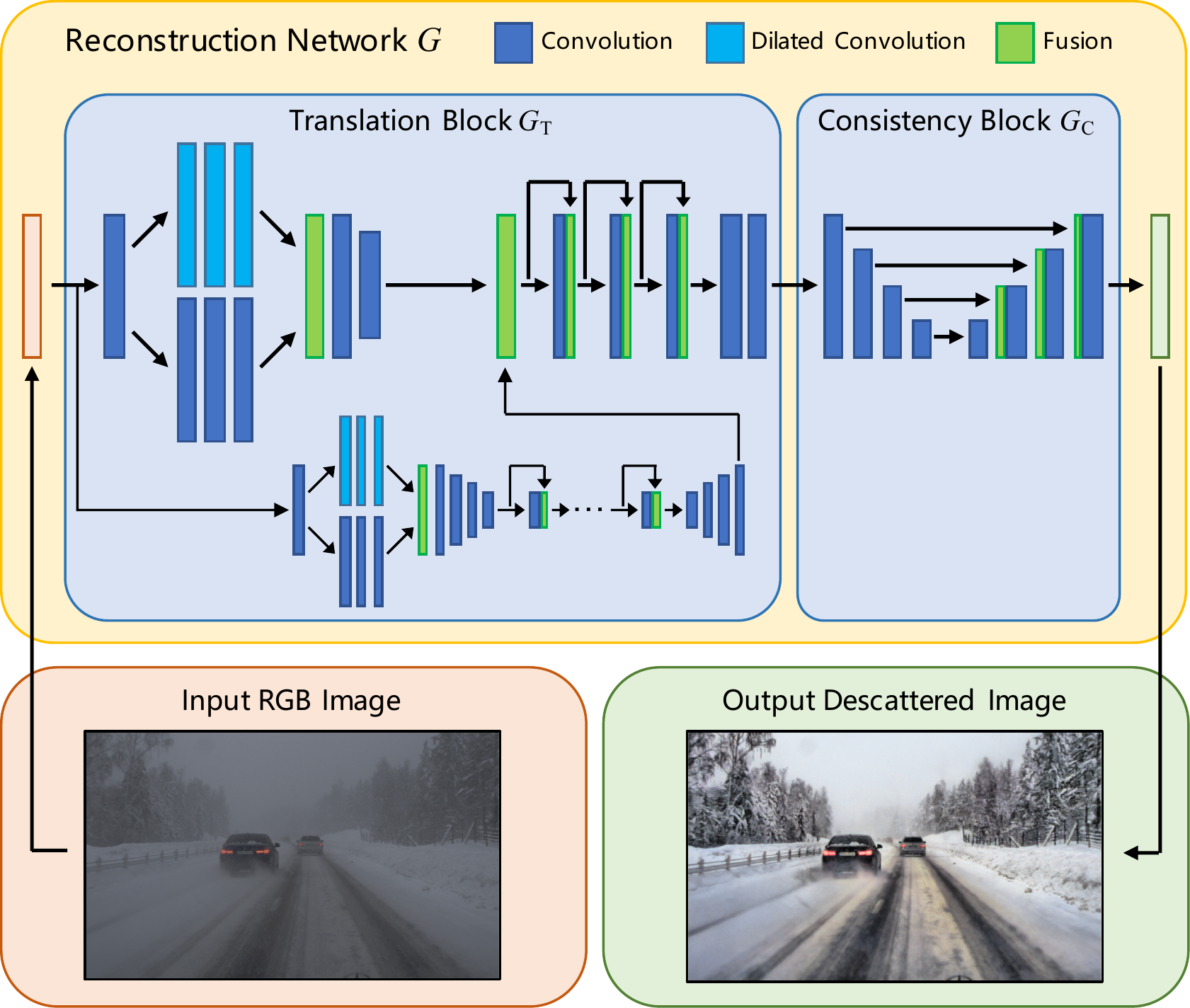}
	\vspace*{-4pt}
	\caption{ZeroScatter generator network architecture. Our generator consists of a translation block that translates raw RGB captures into clear daytime scenes and a consistency block that removes erratic scattering media such as snowflakes.}
	\label{fig:gen_arch}
	\vspace*{-10pt}
\end{figure}

\subsection{Generator Architecture}

Our ZeroScatter generator network is illustrated in Figure~\ref{fig:gen_arch}. The architecture consists of two sequential components: a translation block $G_\text{T}$ that eliminates scattering and performs domain transfer from a raw RGB adverse weather capture into a clear daytime scene, and a consistency block $G_\text{C}$ that further refines the translated output by removing stereo and temporal artifacts. Drawing inspiration from recent image translation networks~\cite{Ignatov2020ReplacingMC,Wang2018}, our translation block architecture consists of two streams, one which operates at the full resolution and the other at a lower resolution. To allow the network to better recognize global features, we use an extended encoder with parallel feature extraction streams: one with $3 \times 3$ convolution layers to extract relative local context and one with $5 \times 5$ kernels with a dilation rate of 2 to allow the network to extract greater global context. Our consistency network consumes the output of the translation block and enforces consistency by removing distortions caused by adverse effects such as snowflakes and sensor noise. The architecture follows a U-Net~\cite{ronneberger2015u} structure with 4 downsampling stages.

\section{Unpaired Training Data and Setup}
We train our model using a dataset from Bijelic et al.~\cite{Bijelic_2020_STF}, who captured harsh weather scenarios in over $\SI{10000}{km}$ of driving in northern Europe. Unlike previous works~\cite{Bijelic_2020_STF,gruber2019pixel,Bijelic2019} we also leverage temporal sequences. The dataset we use consists of 12997 video sequences of length $\SI{0.5}{s}$ and acquired at $\SI{20}{Hz}$, resulting in a total of 120000 individual frames. The video sequences allow us to train for weather and sensor degradations that fluctuate over time, such as sensor noise and snowflakes. Please refer to the Supplemental Document for details on dataset distribution, split, and implementation details of the proposed approach.

We train ZeroScatter using Adam~\cite{kingma2014adam} with a learning rate of $5\mathrm{e}{-5}$. After training, we implement the reconstruction network for real-time inference at $20$ FPS using fp16 precision for $768 \times 1280$ resolution images using an NVIDIA GeForce RTX 2080 Ti GPU. This allows for real-time vision and display applications in automotive systems.

\section{Assessment} 

In this section, we validate the proposed method quantitatively and qualitatively. Our quantitative evaluation is performed on two test sets with paired clear reference data: fog chamber measurements, see Supplemental Material, that allow us to assess robustness to adverse weather in controlled fog scenarios, and a synthetic dataset where the scattering media is produced by $F_\text{Syn}$. For additional experimental details and qualitative results on the synthetic dataset, please refer to the Supplemental Document. Before reporting the performance of the proposed method compared to state-of-the-art image reconstruction approaches, we first validate model architecture choices in an ablation study.

\begin{table*}[t]
	\caption{Quantitative ablation study of different network structures and loss combinations on the fog chamber measurements.}
	\vspace{-5pt}
	\label{tab:ablation_eval}
	\small
	\centering
	\newcolumntype{C}[1]{>{\centering\let\newline\\\arraybackslash\hspace{0pt}}m{#1}}
	\setlength{\tabcolsep}{5pt}
	\begin{tabular}{lC{35pt}C{35pt}C{40pt}C{40pt}C{40pt}C{40pt}C{40pt}C{40pt}}
		\toprule
		& $G_\text{T}$ & $G_\text{C}$ & $\mathcal{L}_\text{model}$ & $\mathcal{L}_\text{multi-modal}$ & $\mathcal{L}_\text{consistency}$ & 1 - LPIPS      & PSNR          & SSIM  \\ \midrule 
		ZeroScatter                             & \checkmark   & \checkmark   & \checkmark           & \checkmark        & \checkmark        & \textbf{0.878} & \textbf{18.8} & \textbf{0.695} \\
		Model \& Multi-Modal cue                & \checkmark   & -            & \checkmark           & \checkmark        & -                &      0.875     &     18.5      &  0.685  \\
		Model cue only                          & \checkmark   & -            & \checkmark           & -                 & -                &      0.870     &     17.4      &  0.658   \\
		$G_\text{T}$ only                       & \checkmark   & -            & \checkmark           & \checkmark        & \checkmark       &      0.872     &     16.9      &  0.665   \\
		Encoder-Decoder~\cite{ronneberger2015u} & -            & -            & \checkmark           & \checkmark        & \checkmark       &      0.870     &     16.6      &  0.650   \\  \bottomrule
	\end{tabular}
	\vspace{-2pt}
\end{table*}

\begin{figure}[t!]
	\centering
	\includegraphics[width=0.48\textwidth]{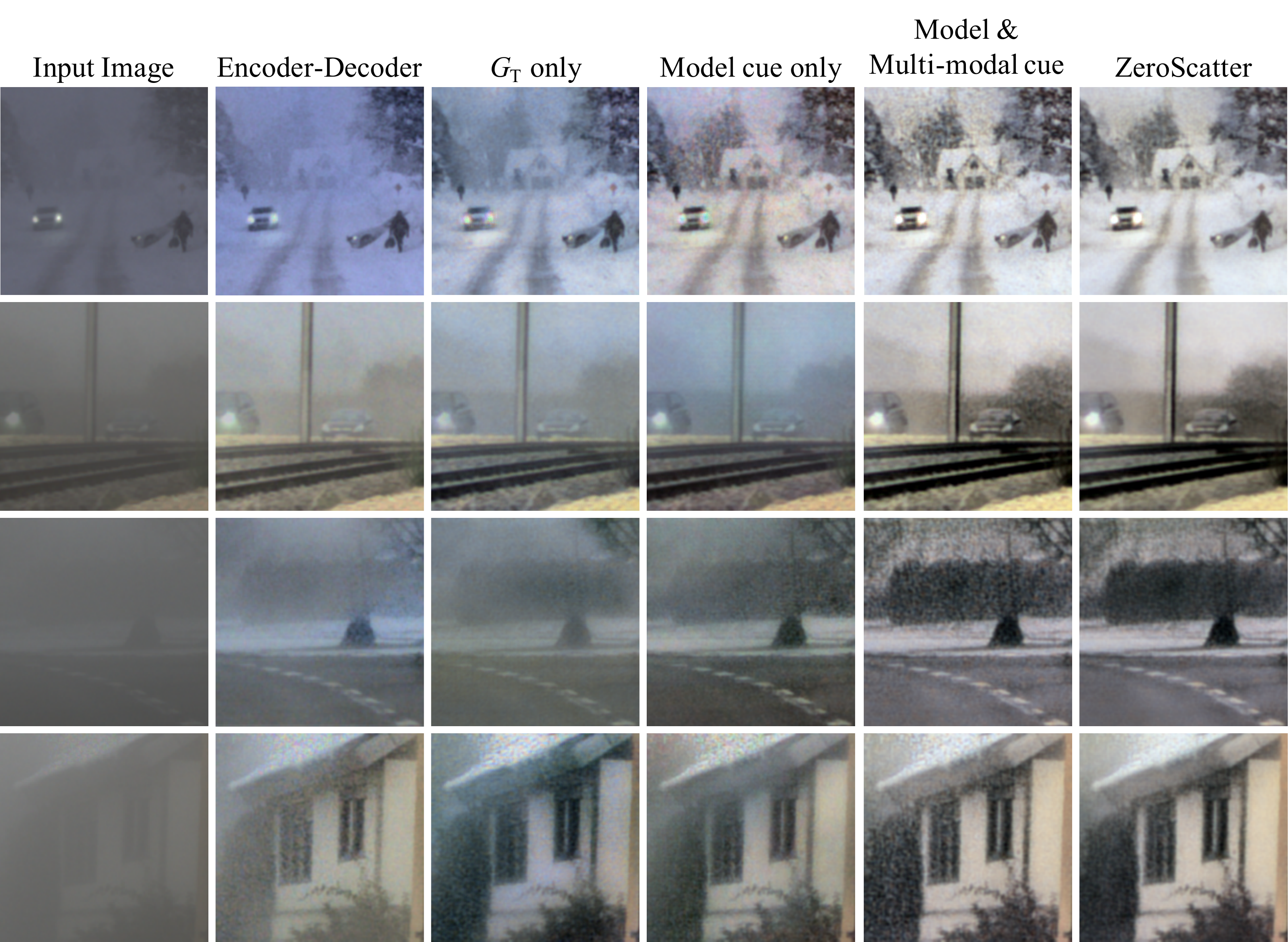}
	\vspace*{-16pt}
	\caption{Ablation study qualitative results on unseen automotive RGB captures. Our sequential architecture and composite loss design enables enhanced contrast at long distances while minimizing snowflakes and sensor noise.}
	\label{fig:ablation}
	\vspace*{-14pt}
\end{figure}

\begin{table*}[t]
\caption{Quantitative evaluation of image descattering methods. We evaluate descattering performance on the synthetic dataset and with controlled fog chamber measurements, see Supplemental Document. We also evaluate object detection performance after applying each descattering method.}
\label{tab:synth_eval}
  \centering
  \vspace{-5pt}
  \small
  \setlength{\tabcolsep}{7pt}
  \begin{tabular}{lccccccccc}
  \toprule
  & \multicolumn{3}{c}{Fog Chamber Measurements} & \multicolumn{3}{c}{Synthetic Dataset}  & \multicolumn{3}{c}{Object Detection} \\
 & 1 - LPIPS      & PSNR          & SSIM            & 1 - LPIPS      & PSNR          & SSIM           & Easy mAP  & Med mAP  & Hard mAP\\ \midrule
\textbf{ZeroScatter} &   \textbf{0.878} & \textbf{18.8} & \textbf{0.695}  & \textbf{0.873} & \textbf{19.2} & \textbf{0.750}  &  91.36      &    \textbf{90.11}     &    \textbf{82.71}\\
  EPDN~\cite{EnhancedPix2Pix}     &  0.844 &  12.7  &  0.565 &  0.840    &  18.4     &   0.715     & \textbf{91.60}  & 88.50 & 80.08  \\
  PFF-Net~\cite{mei2018pffn}    &  0.841  & 15.6  &  0.627 &  0.827    &  18.4     &   0.707     &  91.37  & 89.48 & 81.01    \\
  Bidirectional-FCN~\cite{mondal2018imagedb} &  0.830   & 12.9   &  0.559&  0.847    &  14.4   &   0.673       & 91.21  & 87.11 & 80.94 \\
  DehazeNet~\cite{cai2016dehazenet}    &  0.799  & 9.60  &  0.390  &  0.814    &   13.6    &   0.575     &  91.02  &  85.90 & 80.43   \\
  CyCADA~\cite{hoffman2018cycada}        	   &  0.819   & 13.1   &  0.506  &   0.808    &   14.3  &   0.572     & 90.97  & 88.18  & 80.46    \\
  CycleGAN~\cite{zhu2017unpaired}                &  0.779  & 11.7   &  0.505 &   0.794    &   13.7   &   0.578      & 90.99  & 85.56 & 80.15  \\
  ForkGAN~\cite{zheng_2020_ECCV}       		  &  0.718  & 11.6   &  0.374 &  0.720    &  13.8     & 0.383        & 87.81   &  84.53 & 78.71         \\
  $F_\text{Proc}$~\cite{Clahe}                &  0.852      &       16.0      &    0.607   &  0.851     &  14.4    &  0.678     &  88.59  & 86.95 &   80.93    \\
  Input Image                                  &  0.812  &  14.3  &   0.517  &   0.753    &  13.4    &   0.492    & 90.50  & 86.50 & 80.91 \\ \bottomrule
  \end{tabular}
  \vspace{-8pt}
\end{table*}

\begin{figure*}[!t]
	\centering
	\includegraphics[width=\textwidth]{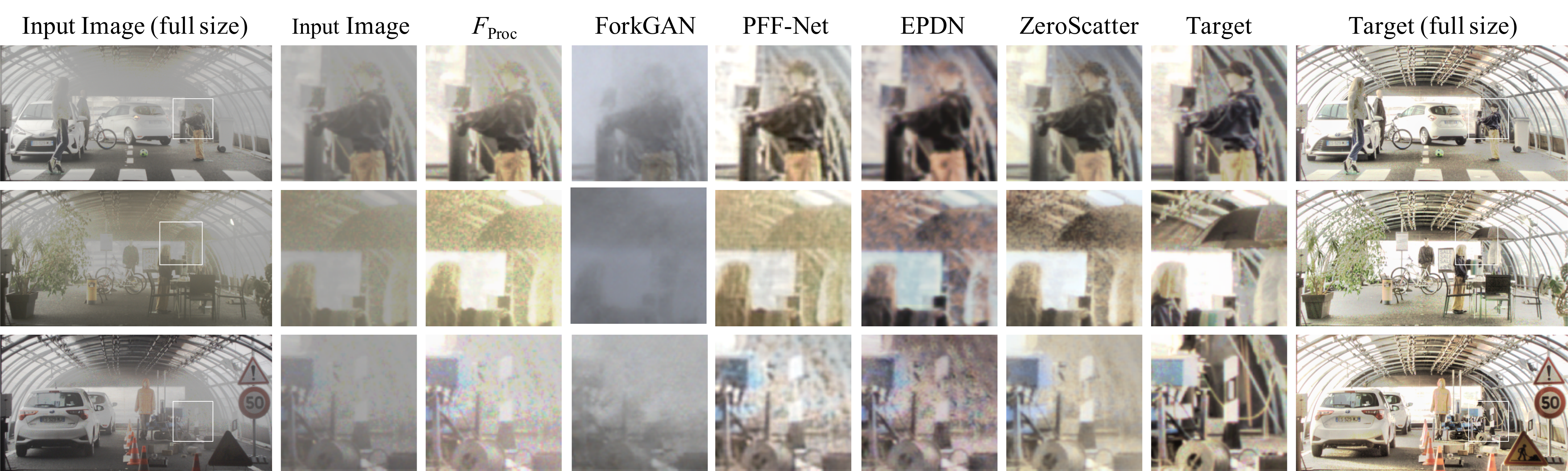}
	\vspace*{-16pt}
	\caption{Qualitative performance comparison on controlled fog chamber measurements, see text. The proposed method significantly reduces scattering media present in the scene and most closely resembles the processed daytime target image.}
	\label{fig:qual_sim}
	\vspace*{-10pt}
\end{figure*}

\subsection{Ablation Study}
\label{sec:ablation}
We conduct an ablation study to validate the effectiveness of our network architecture and the benefits from our novel combination of model-based, multi-modal, temporal, and multi-view supervision.  
The quantitative results for fog chamber measurements are shown in Table~\ref{tab:ablation_eval} together with the ablation configurations.
Qualitative results are shown on unseen real-world data are shown in Figure~\ref{fig:ablation}. 

We observed that relying solely on model-based training cues limits the performance on real-world data, as shown by the ``Model cue only'' configuration. The model outputs suffer from reduced contrast and this model is unable to adequately handle spurious sensor noise. ``Model \& Multi-Modal cue''  illustrates how incorporating multi-modal indirect supervision improves performance with better removal of scattering components and increased contrast. Adding the consistency supervision grants us our proposed model ZeroScatter, which has the best descattering performance overall. Temporal and stereo consistency supervision enables effective removal of snowflakes and local fluctuations including sensor noise.

On the architecture side, our ablation study demonstrates the benefits of our sequential architecture. If we applied a standard encoder-decoder architecture~\cite{ronneberger2015u} then minimal descattering is achieved, as shown by the ``Encoder-Decoder'' configuration. We attribute this to the limited receptive field which is unable to robustly recognize and remove adverse weather. Our translation block remedies this by using dilated convolutions to obtain a wider field of view and this results in better descattering as shown by the ``$G_\text{T}$ only'' configuration. However, without the consistency block $G_\text{C}$, the translation block $G_\text{T}$ falls into a local minimum where it avoids descattering.  This is because the presence of some types of adverse weather such as haze can inadvertently increase temporal and stereo consistency by blurring out image details.  As a result, our final network architecture that uses both $G_\text{T}$ and $G_\text{C}$ obtains the best performance across all variants compared in this work.

\begin{figure*}[!t]
	\centering
	\includegraphics[width=\textwidth]{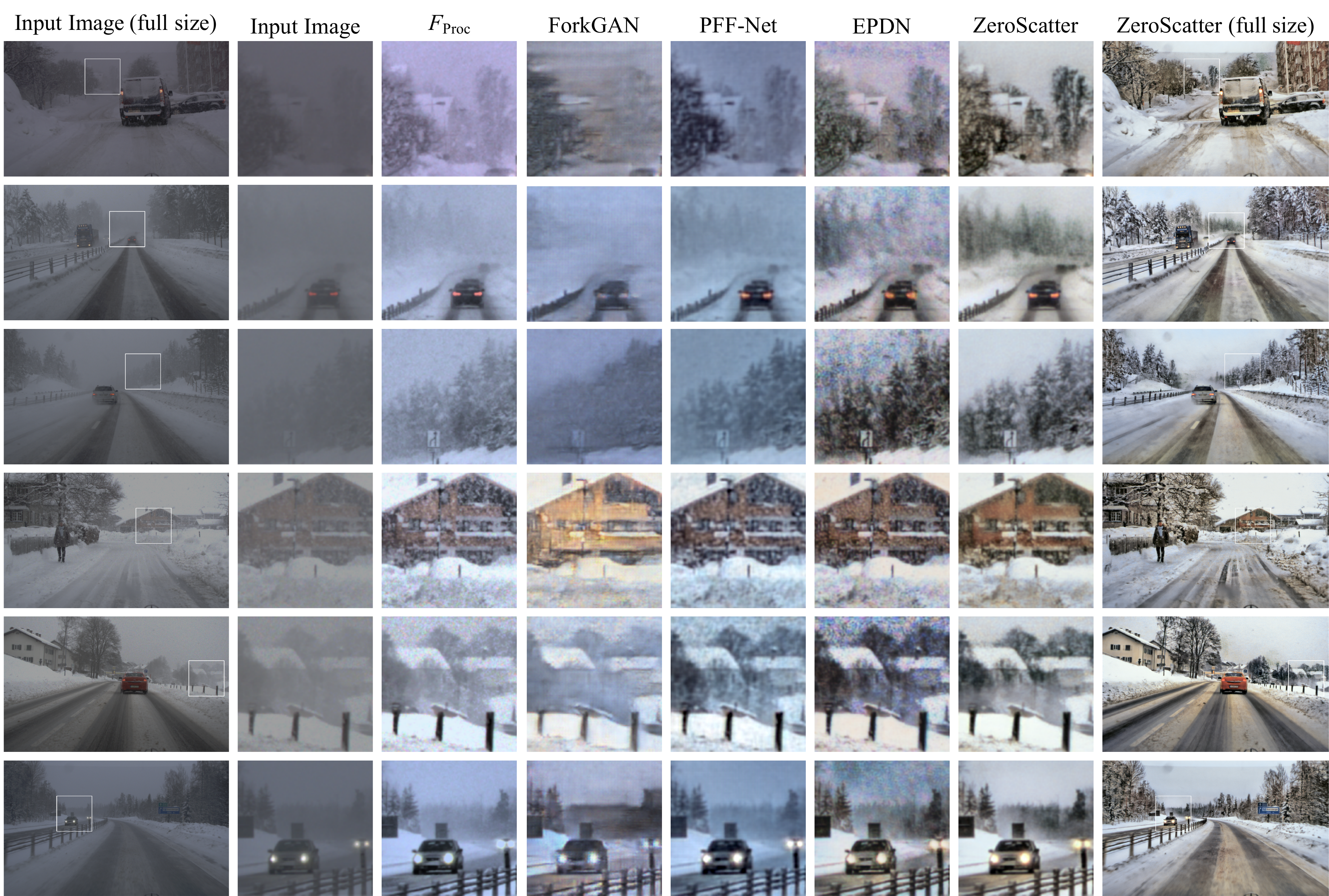}
	\vspace*{-15pt}
	\caption{Real-world data qualitative comparisons. The proposed method significantly reduces scattering present in the scene and reveals object in long distance, such as the house and trees in the top two examples above. Compared to EPDN and PFF-Net, ZeroScatter is able to produce images with better contrast and less noise. ZeroScatter is able to remove snowflakes in the $3^\text{rd}$ and $4^\text{th}$ examples and sensor noise in the $5^\text{th}$ and $6^\text{th}$ examples.}
	\label{fig:qual_results}
	\vspace*{-15pt}
\end{figure*}

\subsection{Controlled Experimental Evaluation}

We compare our work against state-of-the-art image descattering networks \cite{EnhancedPix2Pix, mei2018pffn, mondal2018imagedb, cai2016dehazenet}, image domain transfer networks~\cite{zheng_2020_ECCV,hoffman2018cycada,zhu2017unpaired}, and traditional image refinement techniques~\cite{Clahe}. Quantitative results are shown in Table~\ref{tab:synth_eval} and qualitative results are reported in Figure~\ref{fig:qual_sim}. Please see the Supplemental Document for training details for these baselines and qualitative comparisons against CycleGAN, CyCADA, Bidirectional-FCN, and DehazeNet.

Traditional methods such as CLAHE~\cite{Clahe}  (shown as $F_\text{Proc}$) work well to stylize the image, but fail to remove severe fog and haze in the images. Image domain transfer networks, such as CycleGAN~\cite{zhu2017unpaired}, CyCADA~\cite{hoffman2018cycada} , and ForkGAN~\cite{zheng_2020_ECCV}  obtain better results, but are still unable to recover high-quality images from the degraded input images. Deep learning approaches designed for processing adverse weather such as EPDN~\cite{EnhancedPix2Pix} , PFF-Net~\cite{mei2018pffn} , DehazeNet~\cite{cai2016dehazenet}, and Bidirectional-FCN~\cite{mondal2018imagedb} all perform well on the synthetic dataset, however, these methods are not robust to out of training distribution inputs and consequently fail to generalize to the real-world fog chamber measurements. We attribute this to the inability of these methods to incorporate real-world data into their training scheme. ZeroScatter remedies these limitations and as a result is able to achieve the highest image quality.

\subsection{In-the-Wild Experimental Evaluation}

We showcase the performance of ZeroScatter and the baseline methods on real-world unseen measurements in Figures~\ref{fig:teaser} and \ref{fig:qual_results}. Our high-quality reconstructions shown in these two figures as well as in the Supplemental Document validate the proposed method for diverse real-world scenes. Objects at long distances such as trees, houses, and cars, that have been obscured by adverse weather are revealed by the proposed method. Because the baseline methods do not utilize multi-modal information, their outputs suffer from residual noise and low contrast in the resulting images. Furthermore, without consistency supervision, their processed outputs accentuate sensor noise and fail to remove snowflakes.

\subsection{Descattering for Object Detection}

Furthermore we evaluate whether descattering improves 2D object detection in adverse weather. For this evaluation, we again use real-world adverse weather captures. Ground-truth annotations are performed manually, and difficulty levels are defined based on bounding box height, occlusion level and truncation following~\cite{geiger2012we}. We employ SSD~\cite{liu2016ssd} object detectors with identical architecture that we finetune on the output of each descattering method for a fair comparison.  Quantitative Average Precision (AP) scores are reported in Table~\ref{tab:synth_eval}, qualitative examples and training details are shown in the Supplemental Document. Among all descattering methods, ZeroScatter achieved the highest AP for the medium and hard settings while still maintaining near top performance on the easy setting. We attribute it to ZeroScatter's ability to remove scattering media in adverse conditions which in turn improves object detection through higher confidence detections and bounding box tightness, especially at long distances.

\section{Conclusion}
We introduce ZeroScatter, a novel domain transfer method that maps RGB images captured with strong scattering in adverse weather for removing scattering media from conventional RGB camera captures. We propose a combination of synthetic and real-world data by exploiting model-based, temporal, multi-view, multi-modal, and adversarial training cues. We validate the method by demonstrating that ZeroScatter significantly outperforms approaches both quantitatively in simulation and controlled experimental conditions, and on in-the-wild scenes. Moreover, we validate that removed scattering at long distances with ZeroScatter also enables state-of-the-art object detection results in harsh weather. In the future, we anticipate that ZeroScatter will not only allow human drivers and detectors to see in harsh weather but also assist human annotators for adverse weather scenes, overcoming the fundamental data bias in these scenarios. We envision the proposed training method as a basic building block for vision systems beyond imaging and object detection, especially for autonomous driving and robotics.

\section*{Acknowledgement}
Felix Heide was supported by NSF CAREER Award (2047359) and a Sony Faculty Innovation Award. This research received funding from the Ministry for Economic Affairs and Energy within the project ``VVM – Verification and Validation Methods for Automated Vehicles Level 4 and 5'' (19A19002G). The authors would also like to acknowledge Klaus Dietmayer from University of Ulm.
\clearpage
{\small
\bibliographystyle{ieee}
\bibliography{bib}
}

\end{document}